\documentclass{article}

\usepackage[preprint]{neurips_2023}

\usepackage[utf8]{inputenc} 
\usepackage[T1]{fontenc}    
\usepackage{hyperref}       
\usepackage{url}            
\usepackage{booktabs}       
\usepackage{amsfonts}       
\usepackage{nicefrac}       
\usepackage{microtype}      
\usepackage{xcolor}         
\usepackage{amsmath}         
\usepackage{algorithm2e}
\usepackage{graphicx}
\usepackage{float}

\title{Formatting Instructions For NeurIPS 2023}

\author{%
  Kyle Luther\\
  Princeton Neuroscience Institute\\
  \And
  H. Sebastian Seung \\
  Princeton Neuroscience Institute \\
  Princeton Department of Computer Science \\
  \texttt{sseung@princeton.edu}
}

\begin{document}

\title{DDGM: Solving inverse problems by Diffusive Denoising of Gradient-based Minimization}
\maketitle

\begin{abstract}
    Inverse problems generally require a regularizer or prior for a good solution. A recent trend is to train a convolutional net to denoise images, and use this net as a prior when solving the inverse problem. Several proposals depend on a singular value decomposition of the forward operator, and several others backpropagate through the denoising net at runtime. Here we propose a simpler approach that combines the traditional gradient-based minimization of reconstruction error with denoising. Noise is also added at each step, so the iterative dynamics resembles a Langevin or diffusion process. Both the level of added noise and the size of the denoising step decay exponentially with time. We apply our method to the problem of tomographic reconstruction from electron micrographs acquired at multiple tilt angles. With empirical studies using simulated tilt views, we find parameter settings for our method that produce good results. We show that high accuracy can be achieved with as few as 50 denoising steps. We also compare with DDRM and DPS, more complex diffusion methods of the kinds mentioned above. These methods are less accurate (as measured by MSE and SSIM) for our tomography problem, even after the generation hyperparameters are optimized. Finally we extend our method to reconstruction of arbitrary-sized images and show results on $128 \times 1568$ pixel images.
\end{abstract}

\section{Introduction}

A linear inverse problem is defined by a known measurement operator $A$. Given observed data $y$, the goal is to recover $x$ by ``explaining'' the data, $Ax \approx y$. Traditionally one minimizes the reconstruction error $\Vert Ax - y\Vert^2$, often by some kind of gradient descent. When the condition number of $A$ is large, the inverse problem is said to be ``ill-posed.'' The true minimum of the reconstruction error is a bad solution because it tends to amplify noise. Better results can often be obtained by early stopping of the gradient descent. Another possibility is to formulate a prior probability distribution for $x$, and find the best $x$ by maximizing the posterior probability, treating the reconstruction error as a log likelihood.

Recently, it has been shown that neural nets trained to denoise images can be incredibly successful at generating images when used in a diffusion process \cite{song2020score, ho2020denoising}. Another exciting application of these denoising nets would be as priors for solving inverse problems. Although we have no direct access to the $x$ that gave rise to the data $y$, we assume that we have access to images that are statistically like $x$, i.e., samples from the prior probability distribution $P(x)$ are available. If a net is trained to denoise these samples, it effectively learns something about the prior distribution, and should be helpful for reconstructing the unknown $x$ that gave rise to the data $y$.

We propose a simple method of doing this. The method augments classical gradient-based minimization of the reconstruction error with denoising by the pretrained net. The only perhaps nonintuitive aspect of our method is that noise is also added back in before subsequent denoising.


As far as we know, our simple method is novel. Unlike \cite{kadkhodaie2021stochastic,kawar2021snips,kawar2022denoising, song2023pseudoinverseguided,song2022solving}, our method does not require a singular value decomposition (SVD) to run. Unlike \cite{chung2022improving, chung2023diffusion, graikos2022diffusion} our method does not require backpropagating through the denoiser. And finally, unlike \cite{jalal2021robust} and the previous methods our method does not couple the number of gradient updates to the number of denoiser updates. We'll see that we require an order of magnitude fewer denoiser updates than gradient updates, so our method is fast. We also show that accuracy is also superior, when measured by standard metrics such as MSE or SSIM.


\begin{figure}[H]
    \centering
    \includegraphics[width=\linewidth]{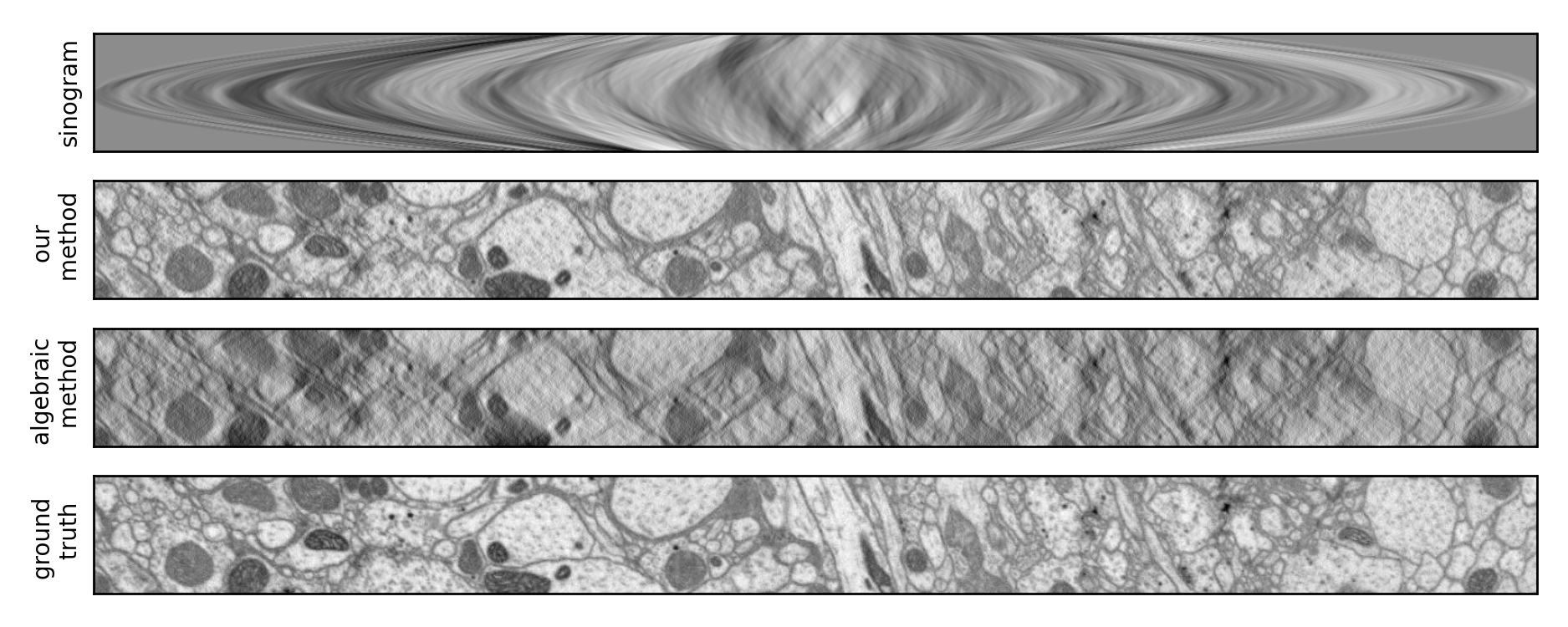}
    \caption{Tomographic reconstruction of a 2D image from 1D projections. (bottom) a 2D image of brain tissue, $128 \times 1568$ pixels at 16 nm resolution
    (top) A series of 1D tilt views obtained by rotating the 2D image by angles in the range $\pm 60^\circ$, and projecting onto a horizontal line. Each row of this ``sinogram'' is one tilt view. (second from bottom) A simple algebraic reconstruction looks ``washed out,'' especially at horizontally oriented image features. (second from top) Many of these features are recovered by our diffusion method.}
    \label{fig:overview}
\end{figure}

We compare our method to denoising diffusion restoration models (DDRM) and a variant of diffusion posterior sampling (DPS) on the inverse problem of tomographic reconstruction from tilt series electron micrographs, a popular technique in biological imaging \cite{mastronarde2017automated}. 2D images of a specimen with a slab geometry are acquired at multiple tilt angles, and a 3D image is inferred by solving the linear inverse problem (Fig. 1). The problem is highly ill-posed, because the angles span a limited range, typically $(-60^{\circ}, +60^{\circ})$. For simplicity, we will study the problem of reconstructing a 2D image from 1D projections. The generalization to reconstructing a 3D image from 2D projections is conceptually straightforward and will be discussed elsewhere, because implementing a 3D denoising net is somewhat more involved.

The generalization of our method to other kinds of inverse problems is very natural. However we have not explored such applications here, because each inverse problem will require some tuning of annealing schedules. This seems to be the case for diffusion methods more generally. We had to extensively tune other methods to achieve performance even competitive with a traditional (non-neural) gradient descent method.

On another note, electron micrographs can be extremely large, and this is the case for biomedical images more generally. Another contribution of this paper is a novel patch-based diffusion process that enables denoisers trained on small patches to handle arbitrarily large images, either for generating image or solving inverse problems. In related work, GANs were used to synthesize images resembling electron micrographs of brain tissue \cite{jain2017adversarial}, but the GANs were not applied to inverse problems.

\section{Diffusive Denoising of Gradient-based Minimization (DDGM)}
\SetKwComment{Comment}{/* }{ */}
\SetKwInOut{Parameter}{Parameter}

\newlength{\commentWidth}
\setlength{\commentWidth}{7cm}
\newcommand{\atcp}[1]{\tcp*[r]{\makebox[\commentWidth]{#1\hfill}}}

\begin{algorithm}
\caption{Diffusive denoising of gradient-based minimization, a method using a denoising net as a prior for solving linear inverse problems (find $x$ such that $Ax \approx y$)}\label{alg:agd}
\KwData{$y \in \mathbb{R}^{m}$ the measured data}
\Parameter{$A \in \mathbb{R}^{m \times n}$ measurement matrix}
\Parameter{$K$ number of gradient iterations per denoising step}
\Parameter{$\lambda$ gradient descent step size}
\Parameter{$\sigma_1, \sigma_N, N$ noise schedule parameters, initial noise, final noise, number of steps}
\Parameter{$\epsilon_\theta$ a denoising network which predicts the normalized noise content of image}

\KwResult{$x \in \mathbb{R}^n$ the reconstruction}
$x \gets 0$\;
$\{\sigma_n\}_{n=1}^N = \{\sigma_1 \cdot (\frac{\sigma_N}{\sigma_1})^{\frac{n-1}{N-1}} \}_{n=1}^N $ \Comment{exponentially decaying noise levels}
\For{$n=1$ \KwTo $N$}{
    \For{$k=1$ \KwTo $K$}{
        $x \leftarrow x - \lambda \nabla_x \Vert Ax - y \Vert^2$
    }
    $ x \leftarrow x + \sigma_n \epsilon_n \;\;\; \text{ where } \epsilon_n \sim \mathcal{N}(0,1)$  \\
    $ x \leftarrow x - \sigma_n \epsilon_\theta (x)$
}
\end{algorithm}


We assume that a network $\epsilon_\theta$ has already been trained to denoise images $x$. We discuss the training objective later in Eq. \ref{eq:MSE}. Our diffusion method for inverse problems is given in Algorithm \ref{alg:agd}. We take $K$ gradient descent steps on the reconstruction error $\Vert Ax-y\Vert^2$. Then we add noise to $x$. Then we denoise $x$ using the net. This process is repeated $N$ times with a noise level that decays exponentially.

We note that the $K$ gradient descent steps on the reconstruction error is essentially a classical algorithm, and by itself already yields some sort of solution. It might be tempting to simply denoise this with our net, but we had little success with this. We speculate this is because the the output of algebraic reconstruction is not simply the real image corrupted by Gaussian noise, which is the only kind of corruption that the net has been trained to remove.

The trick is to \textit{add Gaussian noise} to $x$ before applying the denoising net. If we add enough noise, our denoiser appears to improve $x$ in some ways, at the expense of making it blurry. If this process is repeated with a decaying noise level, we will see that $x$ becomes both sharp and accurately produces features of the true image.

\section{Experiments}
\textbf{Dataset} We downloaded two volumes of size $1k \times 10k \times 1k$ from the center of a publicly available 3D image dataset acquired by FIB-SEM from a fly brain \cite{scheffer2020connectome,xu2020connectome}. We chose the location of these sub-volumes to avoid stitching artifacts that are present in the full dataset. The voxel sizes at MIP-1 resolution are $16 \times 16 \times 16$ nanometers. We used one volume for training, and the other for validation and testing. The dataset was normalized to have zero mean and unit variance computed over all the training set pixels.

\textbf{Training the denoising network} We train a U-Net to denoise $128\times 128$ images corrupted by adding randomly rescaled Gaussian noise $\sigma\epsilon$ to a clean image $x$. The elements of the vector $\epsilon$ are drawn from a Gaussian distribution with zero mean and unit variance. The scalar $\sigma$, or noise level, is chosen from a LogUniform distribution, i.e., $\log \sigma$ is uniformly distributed in the interval $[\log(0.03), \log(30.0)]$. The network output is denoted by $\hat{\epsilon}_{\theta}(x+\sigma \epsilon)$, where $\theta$ are the network parameters and $x+\sigma\epsilon$ is the corrupted image. The network is trained to predict the unscaled noise $\epsilon$, i.e., we minimize the mean squared error
\begin{equation}\label{eq:MSE}
    \nabla_\theta \Vert \epsilon - \hat{\epsilon}_{\theta}(x+\sigma \epsilon) \Vert^2
\end{equation}
using the Adam optimizer with default PyTorch parameters. We train for $380,000$ gradient updates which took 20 hours using 8 NVIDIA 3090 GPUs with a batch size of 64 (8 images per GPU).

Our U-Net style architecture has $128$ feature maps at all 5 levels of resolution, Group Normalization \cite{wu2018group}, residual connections within blocks, and 6 convolutional layers in each block. The net has 8 million parameters in total. 
Following the work of \cite{song2020score}, the net is not conditioned on the noise level, unlike many other models in the diffusion model literature \cite{ho2020denoising}. Rather, a single unconditional net is trained to denoise at any noise level. 

Above the target task was characterized as noise prediction. However, it is more intuitive to flip the sign and think of the target task as denoising. If we regard the output of the net as $-\hat{\epsilon}_{\theta}$, the net is trained to predict a direction in image space that is denoising. Traditionally, a denoising autoencoder is trained to predict a clean image in one step. Our net might be called a residual denoising autoencoder, since it predicts the direction of the difference between the clean and noisy image. This is suitable for the iterative diffusion method that will be introduced below. Note that the net is not trained to predict the magnitude of the denoising step, since the target is the unscaled noise $\epsilon$. Later on, our diffusion procedure will rescale the denoising direction $-\hat{\epsilon}_{\theta}$ appropriately.

Our networks are trained using PyTorch \cite{paszke2019pytorch} and PyTorch Lightning \cite{falcon2019lightning}.


\subsection{Unconditional generation} 
\begin{figure}
    \centering
    \includegraphics[width=\linewidth]{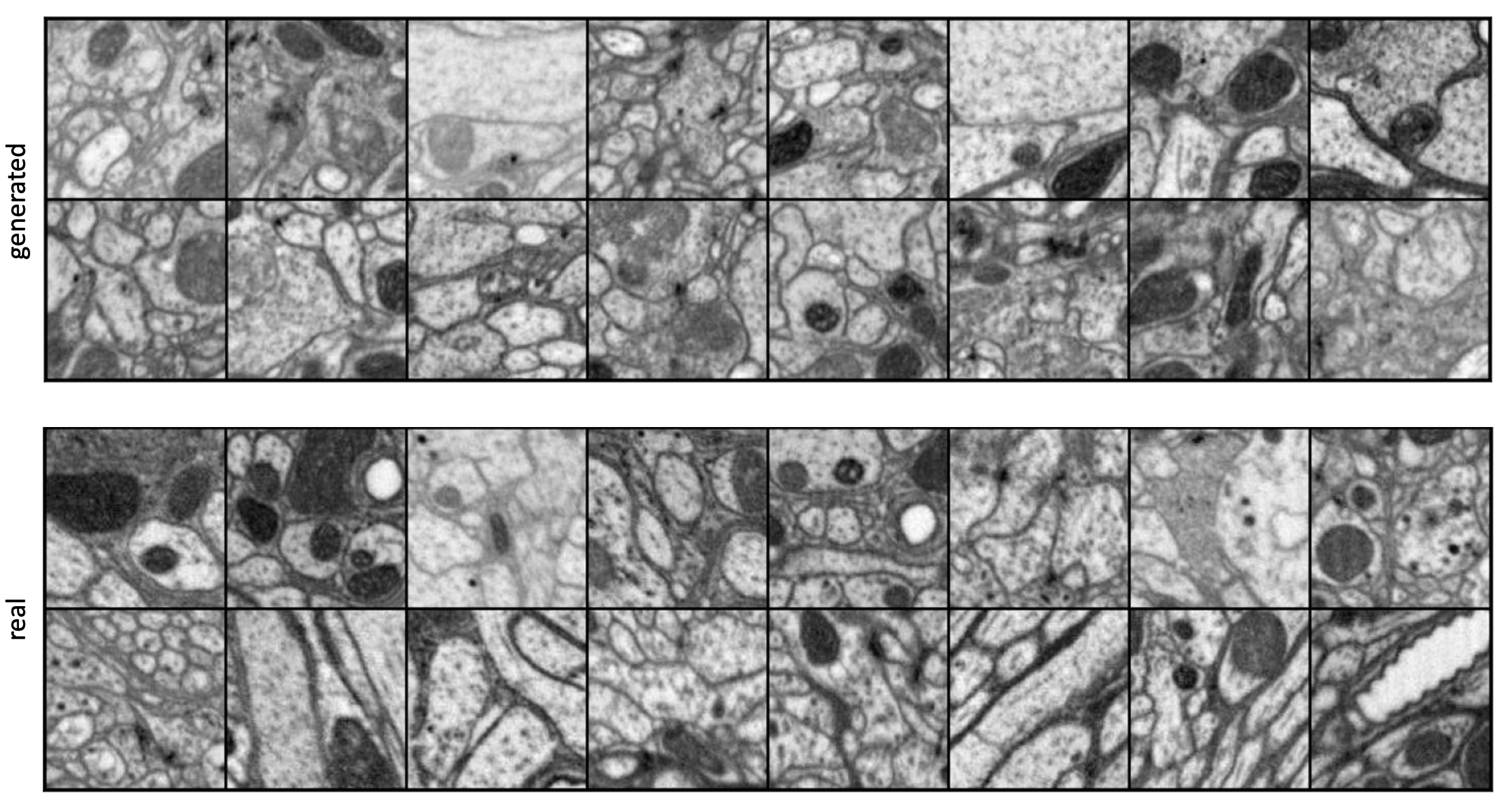}
    \caption{Unconditionally generated samples from our prior vs real samples. Unconditional generation (i.e. not solving an inverse problem) is useful for evaluating the quality of the denoising prior.}
    \label{fig:unconditional}
\end{figure}

Our ultimate goal is to solve an inverse problem, i.e., generate an image that explains the data. However, unconditional generation of images turns out to be invaluable for evaluating the quality of the prior learned by the denoising network, and for adjusting the parameters of the diffusion schedule. We find a simple exponential decay works well enough with 50 diffusion steps. Specifically we initialize $\sigma_1= 30.0$ and $x_1 = \sigma_1 \epsilon_1 $ where $\epsilon_1 \sim \mathcal{N}(0,1)$. We iterate the following for 50 steps to generate images unconditionally:
\begin{equation}
\begin{split}
    \sigma_n = \sigma_1 ((1-\alpha)^2 + \alpha \beta)^{(n-1)/2} \\
    x_{n+1} \leftarrow x_n - \alpha 
    \sigma_n \epsilon_\theta(x_n) + \sqrt{\alpha \beta} \sigma_n \epsilon_n
\end{split}
\label{eqn:diffusion}
\end{equation}
We set $\alpha=0.183$ and $\beta=0.5$ as constants.  This schedule is motivated by the simple exponential-decay schedule proposed in \cite{kadkhodaie2021stochastic} (discussed before their more sophisticated schedule in their Algorithm 1). Our results are shown in \ref{fig:unconditional}. This shows that this denoiser is indeed quite powerful and should be very helpful as a prior for solving inverse problems.

\subsection{Simulated tomographic tilt series}
\label{sec:paramtuning}
We extract random 2D image patches of size $128 \times 128$ from these volumes to train and evaluate our network. We use the Astra Toolbox to simulate 128 uniformly spaced tilt views over the range $(-60^\circ, +60^\circ)$. Each tilt view is a 1D projection of the original image (128 pixels wide). Each of these 128 views are concatenated to form a $128 \times 128$ dimensional data vector $y$ called the sinogram. This is then corrupted with Gaussian noise of magnitude $4.1$ which gives a signal-to-noise ratio of 10 to 1. The tilt views are a linear function of the images.
\begin{equation}
    y = Ax + \sigma_y \cdot \epsilon \;\;\;\;
\end{equation}
In this setting, the number of observed variables (the dimension of $y$) equals the number of variables we are trying to infer (the dimension of $x$) which in this setting is $128^2$. The matrix $A$ is highly ill-conditioned however with over half of the singular values being smaller than $0.1 \times$ the largest singular value of $A$. We will see that simply performing gradient descent on $\Vert Ax - y \Vert^2$ can recover some structure but inevitably misses a significant portion of critical information. 

We use a validation set of 128 images of size $128 \times 128$ to tune the hyperparameters of all reconstruction methods. We report mean squared errors and SSIM metrics on a different test set of 128 images \cite{Wang2004ssim}. We use TorchMetrics to compute SSIM \cite{nicki2022torchmetrics}.

While the test set is small, the standard error on our measurements is still sufficiently low that we can see clear differences between all methods.

\subsection{Using the prior for tomographic reconstructions} 
We run Alg. \ref{alg:agd} on our simulated tilt series. We report quantitative results in Tab. \ref{tab:gaussian}. We report all methods using the absolute best (meaning lowest MSE on a validation set) settings found and the best settings subject to the number of denoiser evaluations being $50$. All qualitative figures shown use the best settings for any number of denoiser evals. For the best settings, we found $\sigma_1=3.0$, $\sigma_N=0.03$, $N=150$, $K=15$ and $\lambda = 9e-5$. For the best settings at 50 denoiser evaluations, we set $\sigma_1=3.0$, $\sigma_N=0.03$, $N=50$. We set $K=25$ and $\lambda = 9e-5$. We found that performance of our method was still quite high at just $50$ denoiser evaluations.

\begin{figure}
    \centering
    \includegraphics[width=\linewidth]{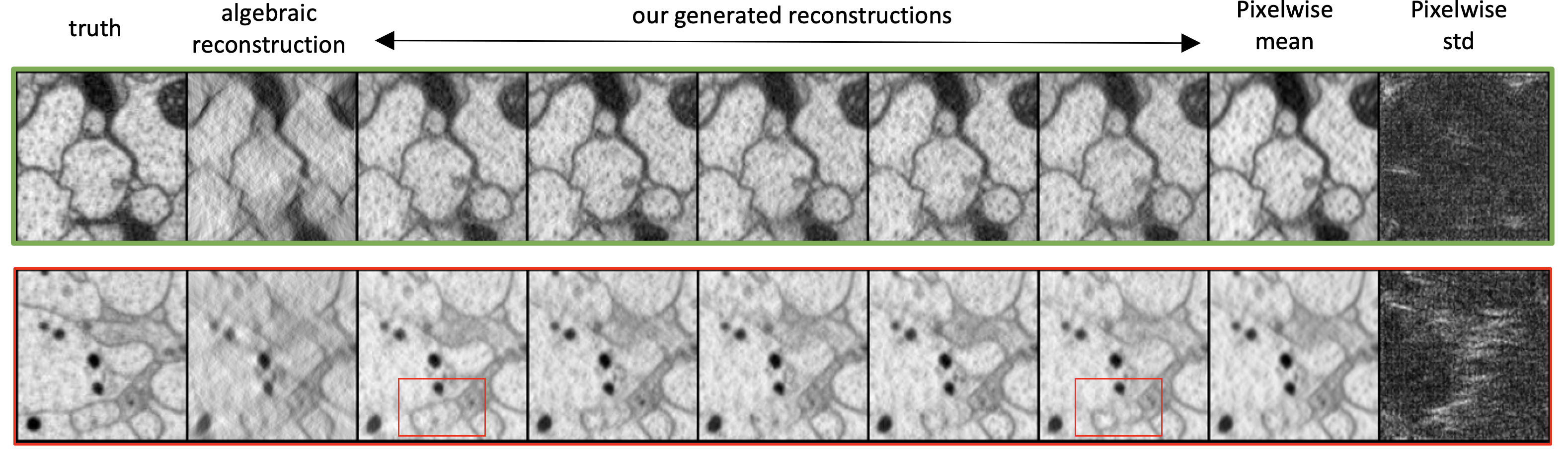}
    \caption{Stochasticity in generated reconstructions. Ideally the method returns virtually the same correct output for all runs (top). Inevitably there will be cases with ambiguities and we observe multiple returned interpretations in this setting (red). }
    \label{fig:variety}
\end{figure}

In \ref{fig:variety} we show two example reconstructions given by our method. Since our method is stochastic, we may end up with different results each time. Ideally these variations would be small, as we would ideally only have one unique solution. For challenging patterns, we occasionally see meaningfully different outputs of the network. In \ref{fig:gaussian} we show single individual reconstructions generated by our method compared to three other reconstruction methods which we'll discuss in the next section. Now we'll discuss how we arrived at this setting of parameters.

\textbf{Step size $\lambda$ for gradient of reconstruction error} We found the largest $\lambda$ for which gradient descent causes the reconstruction error $\Vert Ax -y \Vert^2$ to decrease. For larger $\lambda$, the error explodes. The value of $\lambda$ is held constant for our whole algorithm's process. We kept this value $\lambda = 9e-5$ for all experiments. Future work may explore tuning this parameter as well.

\textbf{Initial noise level $\sigma_1$} The initial noise level we use, $\sigma_1=3.0$, is actually $10\times$ lower than what we used for unconditional generation $\sigma_1=30.0$. Empirically we found that for fixed $N$, lowering the starting $\sigma$ improved reconstruction performance slightly (Appendix). Intuitively, after just a few gradient iterations $\nabla_x \Vert Ax-y\Vert^2$ the reconstruction $x$ already bears some similarity to a real image from the training set. Therefore $x+\sigma \epsilon$ may resemble a clean image + Gaussian noise image for relatively low levels of Gaussian noise. We do not have an explanation for why starting with lower noise levels is actually better for MSE however.

\textbf{Ending noise level $\sigma_N$} We choose the smallest noise level the network was trained on, which in this case was $\sigma_N=0.03$. This is an imperceptibly small level of noise. We did not vary this choice in the experiments.

\textbf{Number of gradient updates $K$ per iteration} This parameter is important. When the number of denoiser evaluations $N=50$, we found the optimal value of $K=25$. When the number of denoiser evaluations $N=150$, we found the optimal value of $K=15$ though the MSE differences were very slight between $K=15$ and $K=25$. Interestingly these values are less than the optimal value of $K=100$ when doing simple algebraic reconstruction (Eq. \ref{eqn:alg}). But we found that the total number of gradient iterations, the product $NK$, was quite large. $NK=1250$ when $N=50$ and $NK=2250$ when $N=150$.

\subsection{Comparisons}
We compare our method to three other methods, DDRM \cite{kawar2022denoising}, a variant of DPS which we call DPS$_*$ \cite{chung2023diffusion}, and a non-neural algebraic reconstruction method. Mean squared error and SSIM are computed between the recovered image $x$ and the ground truth $x_\text{true}$. Results from a test set of 128 images are provided in Tab. \ref{tab:gaussian}. We evaluate the neural methods with two different settings: one where we allow any number of denoiser evaluations and one where we only allow 50 denoiser evaluations (as this setting is much faster).

We find in both cases our method significantly outperforms both DDRM and DPS$_*$ in terms of MSE and SSIM. We find that DPS$_*$ in particular benefits from a large number of denoiser evaluations, but even after 1000 evaluations, its error is far worse than our method with just 50 denoiser evaluations. In this setting of 50 denoiser evaluations, we actually found that the DPS$_*$ method was outperformed even by the simple non-neural algebraic method.

\begin{figure}
    \centering
    \includegraphics[width=\linewidth]{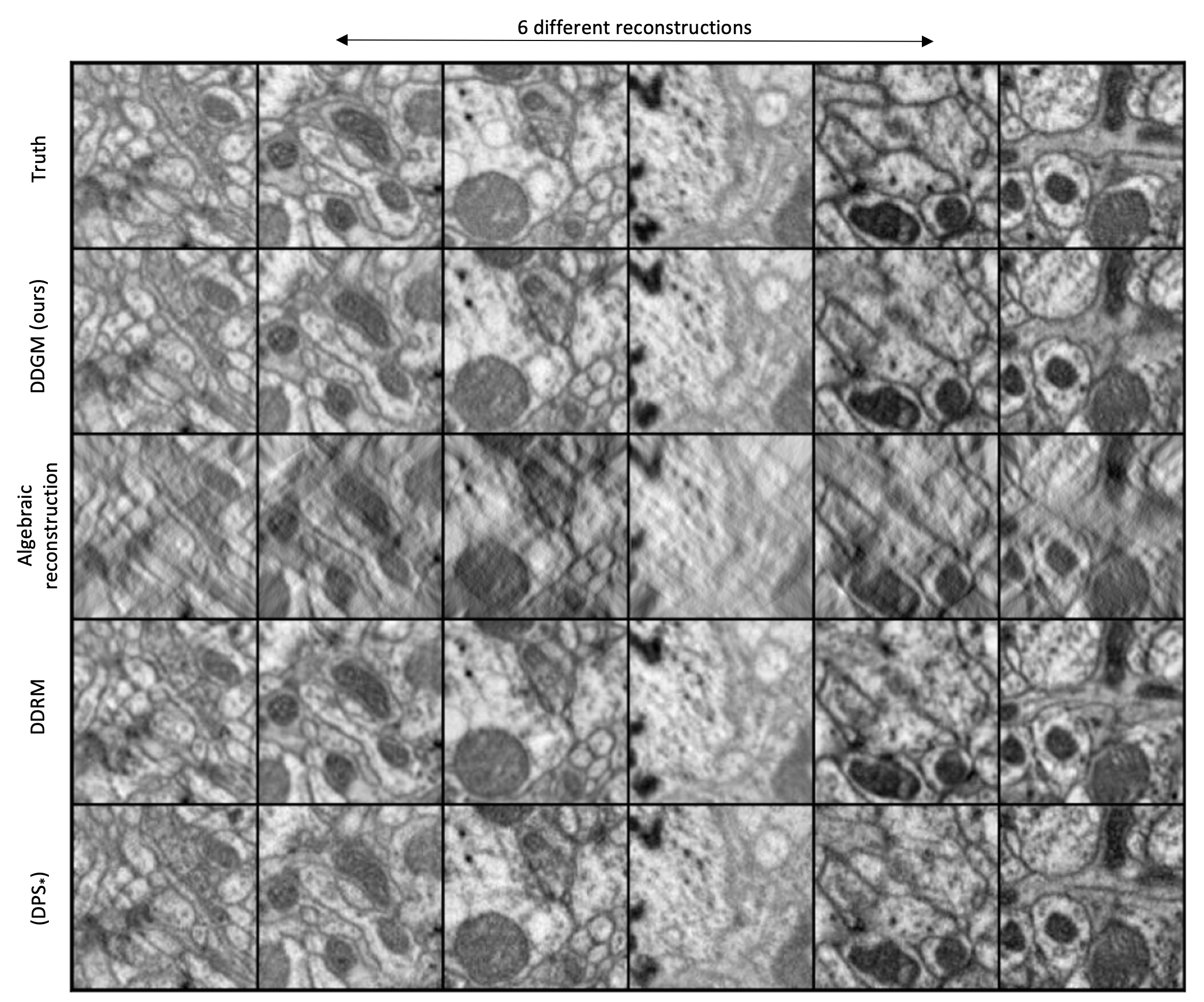}
    \caption{Reconstructions generated via various methods for Gaussian measurement noise. Best viewed zoomed in. DDRM consistently produces blurred responses. DPS$_*$ is competitive by eye with our method, but is inferior by the quantitative measures of Table 1.}
    \label{fig:gaussian}
\end{figure}

\textbf{Algebraic reconstruction} The simplest method does not use a neural network. We just perform $K$ steps of gradient descent on the squared error between predicted tilt views $Ax$ and the measured tilt views $y$
\begin{equation}
    x \leftarrow x - \lambda \nabla_x \Vert Ax - y \Vert^2
\label{eqn:alg}
\end{equation}
Early stopping is used as an implicit regularizer. We initialize $x=0$. We set $\lambda=9e-5$ to be the $\lambda$ which gives rise to the fastest decrease in objective value. This means we have one hyperparameter, the number of gradient steps $K$. We find that $K=100$ gives the lowest MSE between true and generated reconstructions $x$ on our validation set. We show the validation set reconstruction errors as we vary $K$ in the Appendix. We show the test set values for $K=100$ in Table \ref{tab:gaussian} and show reconstructions in Fig. \ref{fig:variety} and Fig. \ref{fig:gaussian}

\textbf{Denoising Diffusion Restoration Models (DDRM)} We refer the reader to Eq. 7 and 8 of \cite{kawar2022denoising} for the full description of the algorithm, which relies on the SVD of the projection operator $A$. We make note that computing the SVD of $A$ is simple enough for $128 \times 128$ images (see Appendix for singular values), but more thought would be required to apply this method to our $128 \times 1568$ pixel images in Fig. 1 due to memory constraints.

This method treats different singular values of the measurement operator differently depending on the level of noise at each step of the diffusion process. We note that the method appears to recommend setting the initial noise level to be larger than the largest non-zero singular value of $A^\dagger$. In our case this would imply setting $\sigma_\text{init} \approx 1 / 10^{-5} = 10^5$. Our network has only been trained on noise levels up to 30 however. To proceed with the DDRM method, we therefore set all singular values of $A$ which are smaller than $1 / 30.0$ to zero and initialize our DDRM diffusion process at $\sigma=30.0$.

We keep $\eta_b = 1.0$ as used in the paper and tune their $\eta$ parameter and the number of diffusion steps $N$. We find that $\eta=1.0$ and $N=10$ provided the minimal mean squared reconstruction error on a validation set of images. We show the validation set reconstruction errors as we vary $\eta$ and $N$ in the Appendix. We show the test set values in Table \ref{tab:gaussian} and show reconstructions in Fig. \ref{fig:gaussian}.

The reconstructions were notably blurry for all settings of parameters we tried (see examples in Fig. 4). This is not inconsistent with the recoveries shown by \cite{kawar2022denoising}. We also observed that unlike many diffusion models, using a surprisingly small number of steps (10 was optimal) gave higher performance than more steps.

\textbf{Diffusion Posterior Sampling (DPS$_*$)} We compare to a variant of the DPS method proposed in \cite{chung2023diffusion}, which we'll call DPS$_*$. We cannot use our networks with their exact diffusion schedule, as they use a diffusion method which learns the variances at each step, and they operate in the "variance preserving" regime (as opposed to the "variance exploding regime" we work in which means the variance of our patterns grows with increasing noise level). However we make several modifications and perform extensive parameter tuning in an attempt to get this method working for our reconstruction problem.

We first apply their key insight, line 7 of their Algorithm 1, to our pre-existing diffusion schedule. Specifically we add a normalized gradient term to the diffusion step of (Eqn. \ref{eqn:diffusion}):

\begin{equation}
\begin{split}
    x_{n+1} = x_n - \alpha \sigma_n \epsilon_\theta(x_n) + \sqrt{\alpha \beta} \sigma_n \epsilon_n - \zeta \frac{\nabla_{x_n} \Vert A (x_n - \sigma_n \epsilon_\theta(x_n)) - y \Vert^2}{\Vert A (x_n - \sigma_n \epsilon_\theta(x_n)) - y \Vert}
\end{split}
\label{eqn:dps_unscaled}
\end{equation}

If we stick with our pre-existing diffusion schedule (so setting $\alpha=0.183$ and $\beta=0.5$ and $N=50$, the schedule used to produce the images in Fig. \ref{fig:unconditional}) then there is only one parameter to tune: $\zeta$. We tune this parameter and show the results in the Appendix. We are unable to set $\zeta$ large enough to make the reconstructions match the data (the MSE is always worse than for the simple algebraic reconstruction method). 

There is a slight technical detail here. We are operating in the variance exploding regime, meaning our $x_n$ are related to their $x'_n$ via $x_n \approx \sqrt{1+\sigma_n^2} x'_n$ so it may be more appropriate to rescale their gradients $\nabla_x$ by $\frac{1}{\sqrt{1+\sigma_n^2}}$. Therefore we also compare a rescaled version of DPS:
\begin{equation}
    x_{n+1} = x_n - \alpha \sigma_n \epsilon_\theta(x_n) + \sqrt{\alpha \beta} \sigma_n \epsilon_n - \zeta \frac{\nabla_{x_n} \Vert A (x_n - \sigma_n \epsilon_\theta(x_n)) - y \Vert^2}{\sqrt{1+\sigma_n^2} \Vert A (x_n - \sigma_n \epsilon_\theta(x_n)) - y \Vert}
\label{eqn:dps_rescaled}
\end{equation}
We find this moderately improves MSEs so we use the rescaled version in the rest of our experiments. We explore what happens when we lower the starting noise level of our diffusion process, lowering $\alpha$ so that $\sigma_N$ is still $0.03$ and $N$ is still 50. We find that lowering the starting noise level helps substantially but does not let us acheive even the MSE given by the classic algebraic reconstruction method. We explain this result as follows: our problem is highly ill-conditioned and if we use our pre-existing diffusion schedule, we only allow 50 gradient steps which simply is not enough for the data to strongly influence the reconstructions.

In their paper, they iterate for 1000 steps, so we modify $N=1000$ and $\alpha$ such that $\sigma_N=0.03$ and perform more experiments, tuning the coefficient $\zeta$. The experiments are rather slow at this point since each image requires backpropagation through our denoiser 1000 times. However we found that $\sigma_1=3.0$ and $\zeta=0.1$ gave optimal performance in this setting.


\begin{table}[]
\begin{center}
\begin{tabular}{ l c c c c c c c c }
Method & \shortstack{requires \\ SVD} & \shortstack{requires \\ $\nabla_x \epsilon_\theta(x)$} & \shortstack{\# grad \\ evals $\downarrow$} & \shortstack{\# net \\ evals $\downarrow$ } & MSE $\downarrow$ & SSIM $\uparrow$ \\ 
DDGM (ours) & no & no & 2250 & 150 & $\mathbf{0.13 \pm 0.004}$ & $\mathbf{0.64 \pm 0.005}$ \\
DDRM & yes & no & N/A & 10 & $0.18 \pm 0.005$ & $0.55 \pm 0.005$ \\
DPS$_*$ & no & yes & 1000 & 1000 & $0.20 \pm 0.005$ & $0.54 \pm 0.004$ \\
\\
DDGM$^{50}$ (ours) & no & no & 1250 & 50 & $0.14 \pm 0.004$ & $0.62 \pm 0.005$ \\
DDRM$^{50}$ & yes & no & N/A & 50 & $0.20 \pm 0.006$ & $0.53 \pm 0.005$ \\
DPS$_*^{50}$ & no & yes & 50 & 50   & $0.35 \pm 0.007$ & $0.34 \pm 0.004$ \\
\\
Algebraic & no & no & 100 & 0 & $0.23 \pm 0.006$ & $0.51 \pm 0.006$ \\
\end{tabular}
\caption{Test set evaluation of various tomographic reconstruction methods. For the neural methods, we show a) the best absolute parameters we found and b) the best parameters with the constraint that we only use $50$ denoiser evaluations. Qualitative outputs are shown in Fig. \ref{fig:gaussian} }
\label{tab:gaussian}
\end{center}
\end{table}

\subsection{Arbitrary-sized image reconstruction}
For this algorithm to have practical utility in connectomics, we must by able to reconstruct arbitrarily large images.  Naively one could try running Algorithm 1 in patches then stitching the outputs together. However, the algorithm is inherently random and it is not obvious how to avoid seams in that scenario. One idea is to attempt to modify the work of \cite{wang2023unlimited} to the setting of inverse problem solving. We take a conceptually simpler approach. Instead we modify the denoiser network itself to run in patches, then we smoothly blend the denoised together. Mathematically, we convolve the denoiser outputs with a 2D bump function. The details are provided in the Appendix. We use this method to produce the $128 \times 1568$ pixel reconstruction shown in Fig \ref{fig:overview}.

\section{Discussion}



\textbf{Future directions} In this paper we focused on a particular inverse problem, limited angle computed tomography. It would be interesting to explore application of our method to other inverse problems, especially non-linear ones. Since we do not require SVD, our method is at least well-equipped in principle to solve non-linear inverse problems such as those considered by the DPS method.

Another line of work should consider annealing the step size or number of gradient steps inside each loop. In our algorithm, the effective strength of the prior decays exponentially over time, while the data-term does not change. Surprisingly we did not need to anneal the data-driven term for our application, but other applications may benefit from such an annealing. Another interesting line of work would be the use of a preconditioner in the gradient updates with the idea of reducing the number of gradient evaluations at each iteration. Currently the gradient updates are the slow component of our algorithm.

\textbf{Limitations} A notable drawback of this method and related methods is the sensitivity to parameters. This work was aided by the fact that we have a ground truth by which we could tune the parameters. However, in the real world, one will typically use tomography to infer the 3D structure of an object that no other method can. This means there is no ground truth on which to tune the parameters, or more generally, evaluate the method. Another limitation regards our evaluation method. We have relied on MSE and SSIM, but these might encourage blurry reconstructions in uncertain image regions. Future evaluations should explore additional quantitative metrics.

\textbf{Potential negative impacts} One concerning outcome of this line of work is the tendency of the networks to hallucinate or eliminate real biological structures. We have observed that the reconstructions usually look very realistic, even when they are incorrect. For scientific applications, such hallucinations can be very concerning. One must take great care to validate any systems that derive scientific results from methods such as ours which use powerful priors to guide data-driven reconstructions.

\bibliographystyle{plainnat}

\appendix
\section{Dataset details and splits}
We download data using the Cloud Volume Python client \cite{silversmith2021cloudvolume} to access to the Janelia Fly Hemibrain dataset \cite{xu2020connectome,scheffer2020connectome}. For the training set, we download a contiguous 10 gigavoxel volume at MIP-1 resolution from corner (x,y,z) = (10750, 6500, 9000) to (x,y,z) = (11750,16500,10000). For the validation and test sets, we extract randomly located patches from a contiguous volume that extends from corner (x,y,z)=(12250,6500,9000) to (x,yz)=(13250,16500,10000).

The whole volume from which these subvolumes were downloaded can be viewed interactively in 3D by visiting the following Neuroglancer link
\url{https://hemibrain-dot-neuroglancer-demo.appspot.com/#!%7B%22dimensions%22:%7B%22x%22:%5B8e-9%2C%22m%22%5D%2C%22y%22:%5B8e-9%2C%22m%22%5D%2C%22z%22:%5B8e-9%2C%22m%22%5D%7D%2C%22position%22:%5B19030.537109375%2C21620.150390625%2C18099.5%5D%2C%22crossSectionScale%22:2.261148454279227%2C%22crossSectionDepth%22:-37.62185354999912%2C%22projectionScale%22:109219.18067006872%2C%22layers%22:%5B%7B%22type%22:%22image%22%2C%22source%22:%22precomputed://gs://neuroglancer-janelia-flyem-hemibrain/emdata/clahe_yz/jpeg%22%2C%22tab%22:%22source%22%2C%22name%22:%22emdata%22%7D%5D%2C%22showSlices%22:false%2C%22selectedLayer%22:%7B%22size%22:290%7D%2C%22layout%22:%22xy-3d%22%7D}


\section{More unconditional generations}
\begin{figure}[H]
    \centering
    \includegraphics[width=\linewidth]{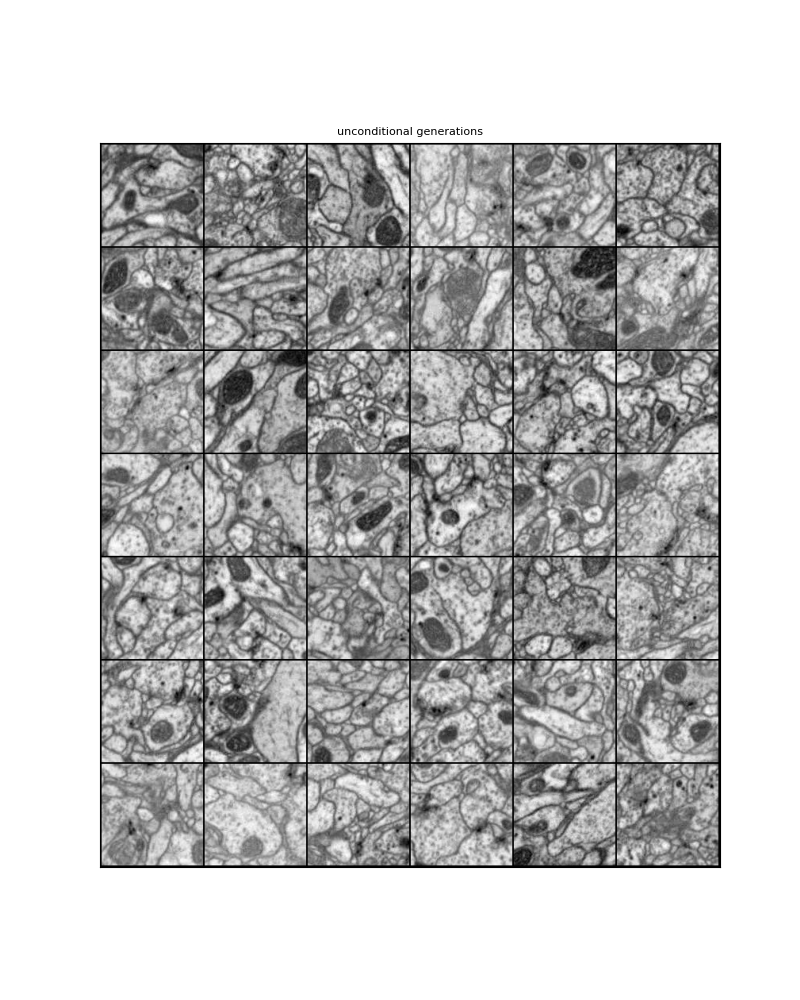}
    \caption{More unconditional generations with our diffusion model. We generate these images using the algorithm and hyperparameters described in Section 3.3}
    \label{fig:my_label}
\end{figure}

\section{More reconstructions from our method}
\begin{figure}[H]
    \centering
    \includegraphics[width=\linewidth]{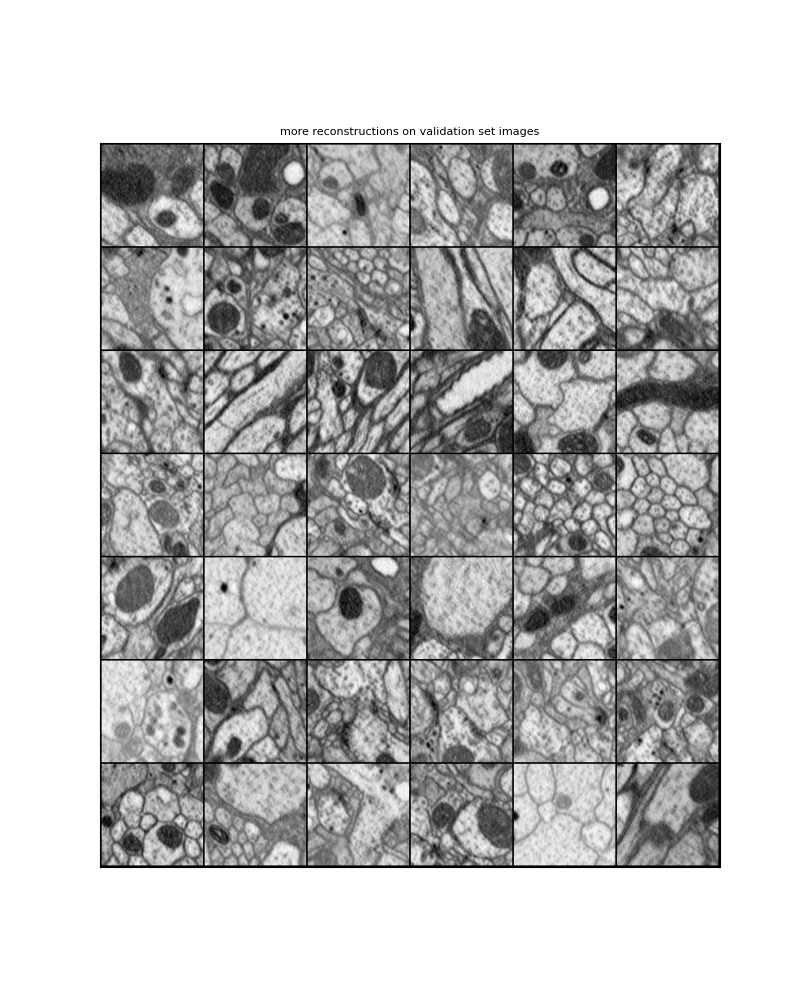}
    \caption{More tomographic reconstructions on validation set images from our model. Tilt views were simulated as described in Section 3.2. We generate these reconstructions using our method with the best-performing hyperparameters described in Table 1 of the paper.}
    \label{fig:more_reconstructions}
\end{figure}

\begin{figure}[H]
    \centering
    \includegraphics[width=\linewidth]{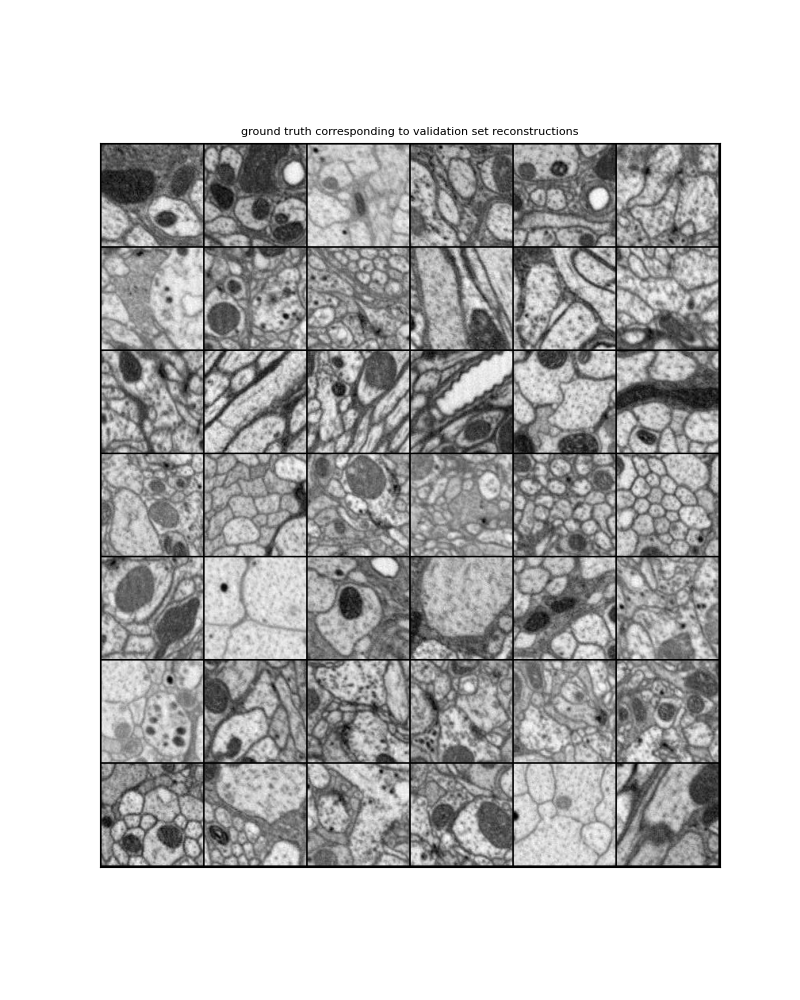}
    \caption{Ground truth corresponding to reconstructions from Fig. \ref{fig:more_reconstructions}}
    \label{fig:my_label}
\end{figure}

\section{Singular values of the projection matrix}
\begin{figure}[H]
    \centering
    \includegraphics[width=\linewidth]{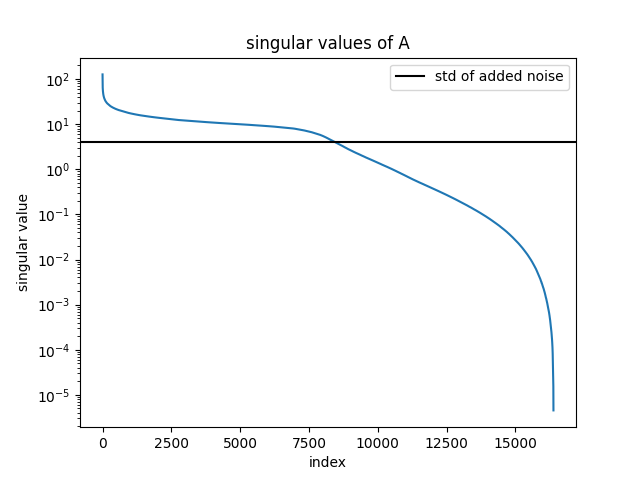}
    \caption{Singular values of the matrix $A$ in the equation $y=Ax+\sigma_y \epsilon$. This matrix A implements the forward projection operator, returning projections of the images from the angular range ($-60^\circ,+60^\circ$). We can see that this matrix is highly ill-conditioned, with singular values spanning a range from $10^2$ down to $10^-5$s.}
    \label{fig:my_label}
\end{figure}

\section{Parameter Tuning}

\subsection{Diffusion Denoising of Gradient Minimization}
We compute performance of our method on the validation set of 128 images as we modify various hyperparameters.
\begin{figure}[H]
    \centering
    \includegraphics[width=\linewidth]{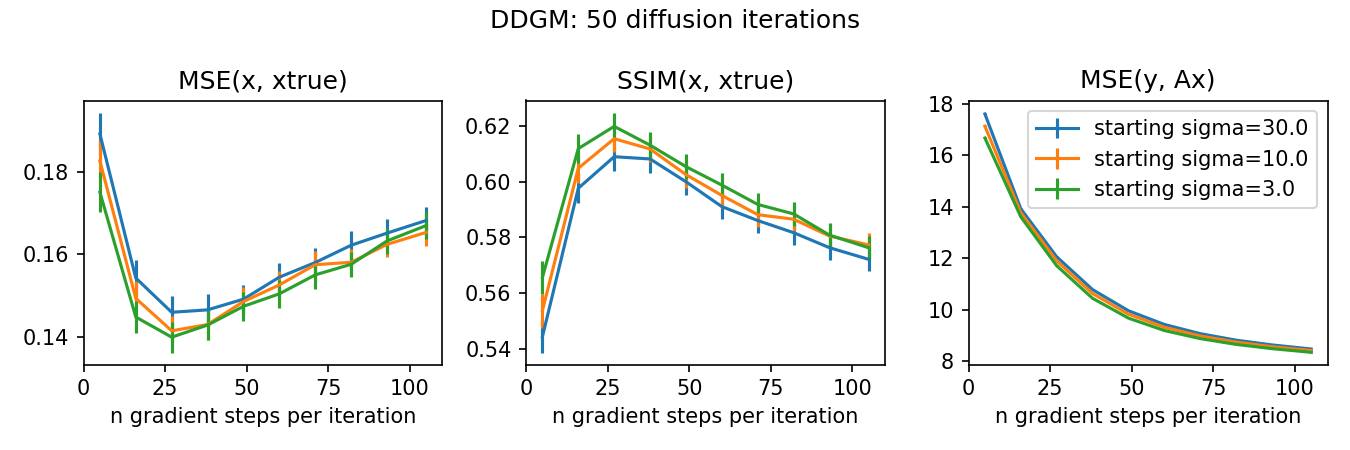}
    \caption{Tuning $K$=number of gradient steps per iteration and starting $\sigma$ with 50 diffusion steps, $\sigma_N=0.03$ (the final noise level). We set $\lambda=9e^{-5}$. The MSE between the predicted measurements $Ax$ and the observed data $y$ decreases monotically as $K$ increases. However, the more important metrics, the MSE between reconstruction $x$ and the ground truth $x_{true}$ reaches its minimum at $K=25$. Similarly the SSIM between reconstruction $x$ and the ground truth $x_{true}$ reaches its maximum at $K=25$.}
    \label{fig:ddgm_50}
\end{figure}

\begin{figure}[H]
    \centering
    \includegraphics[width=\linewidth]{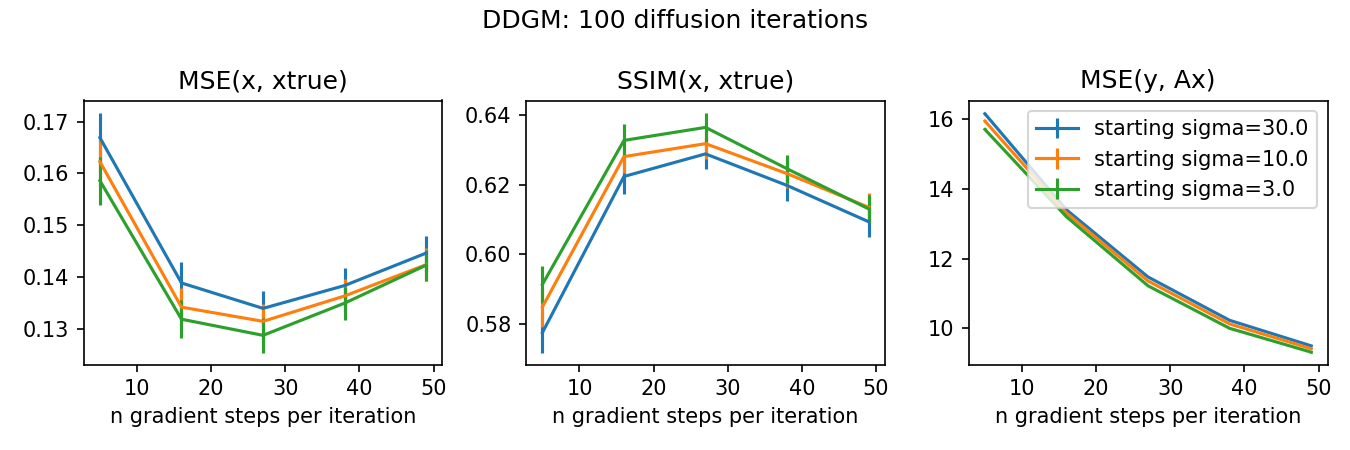}
    \caption{Tuning $K$=number of gradient steps per iteration and starting $\sigma$ with 100 diffusion steps, $\sigma_N=0.03$ (the final noise level). We set $\lambda=9e^{-5}$.}
    \label{fig:my_label}
\end{figure}

\begin{figure}[H]
    \centering
    \includegraphics[width=\linewidth]{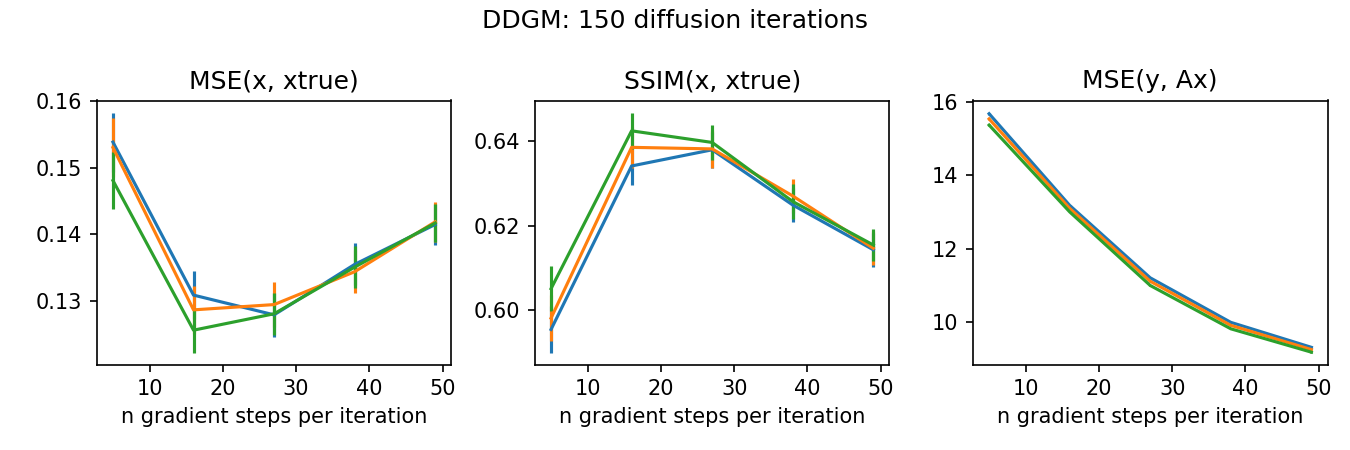}
    \caption{Tuning $K$=number of gradient steps per iteration and starting $\sigma$ with 150 diffusion steps, $\sigma_N=0.03$ (the final noise level). We set $\lambda=9e^{-5}$. Using 150 diffusion iterations (compared to 50 as in Fig \ref{fig:ddgm_50} improves MSE and SSIM between reconstruction $x$ and ground truth $x_{true}$ slightly. Notably the MSE between the predicted measurements $Ax$ and the observed data $y$ (right figure) is higher in this setting than when the number of diffusion iterations is 50.}
    \label{fig:my_label}
\end{figure}

\subsection{Algebraic Reconstruction}

\begin{figure}[H]
    \centering
    \includegraphics[width=\linewidth]{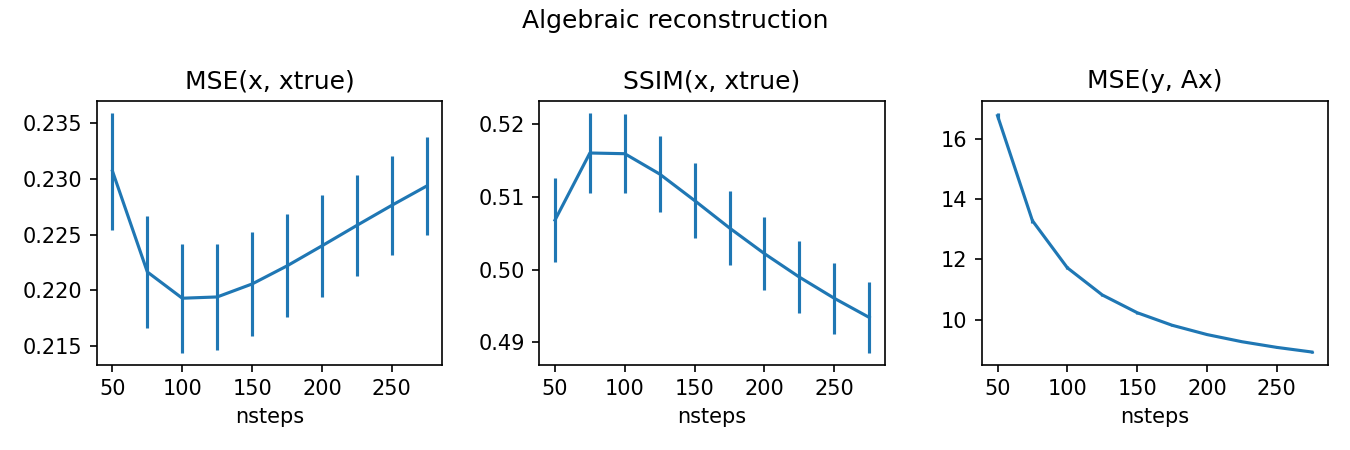}
    \caption{Vary $K$=number of gradient steps taken. The only other hyperparameter is $\lambda=9e^{-5}$.}
    \label{fig:my_label}
\end{figure}

\subsection{Denoising Diffusion Restoration Models}
\begin{figure}[H]
    \centering
    \includegraphics[width=\linewidth]{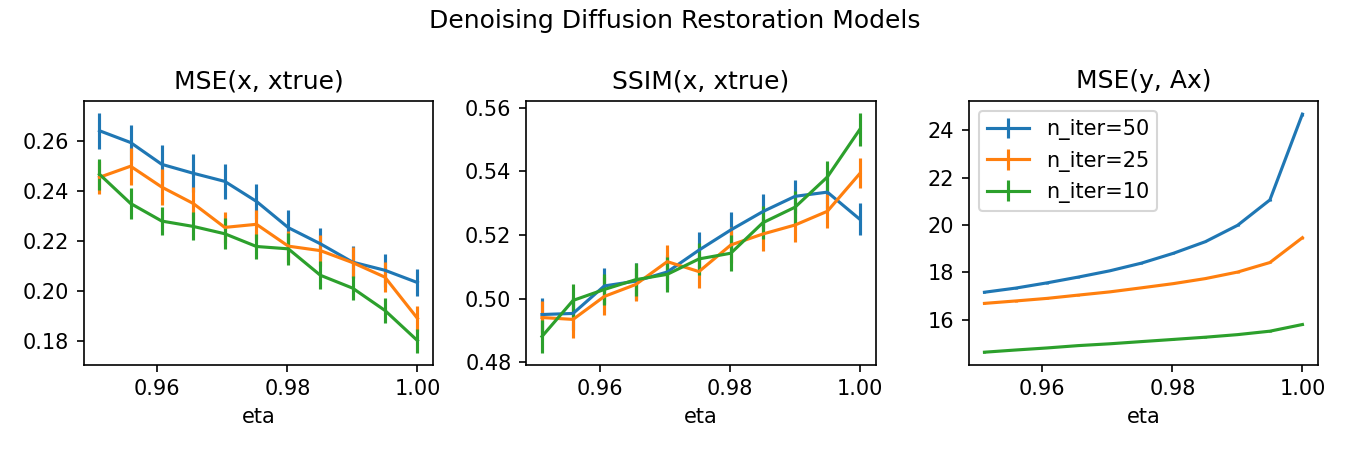}
    \caption{Varying $\eta$ and number of diffusion iterations for DDRM. We also tried niter=5 but the performance was substantially worse than niter=10 and does not fit on these charts.  }
    \label{fig:my_label}
\end{figure}

\subsection{Diffusion Posterior Sampling}
Besides the hyperparameter $\zeta$ governing the gradient step sizes, we find that DPS has an extreme sensitivity to the details of the noise schedule. In particular both the number of diffusion steps and the precise noise levels used have a large impact on the ultimate reconstruction performance. This makes somes sense as the number of gradient steps is tied to the number of diffusion steps in this algorithm. More performance could perhaps be achieved by a more extensive grid search over diffusion schedule parameters, but even evaluating a singular parameter configuration with 1000 diffusion steps (a single point in Fig. \ref{fig:dps_niter1000}) takes over 1 hour so performing a grid search would require significant time investment. Furthermore, we struggled to find any setting of parameter
\begin{figure}[H]
    \centering
    \includegraphics[width=\linewidth]{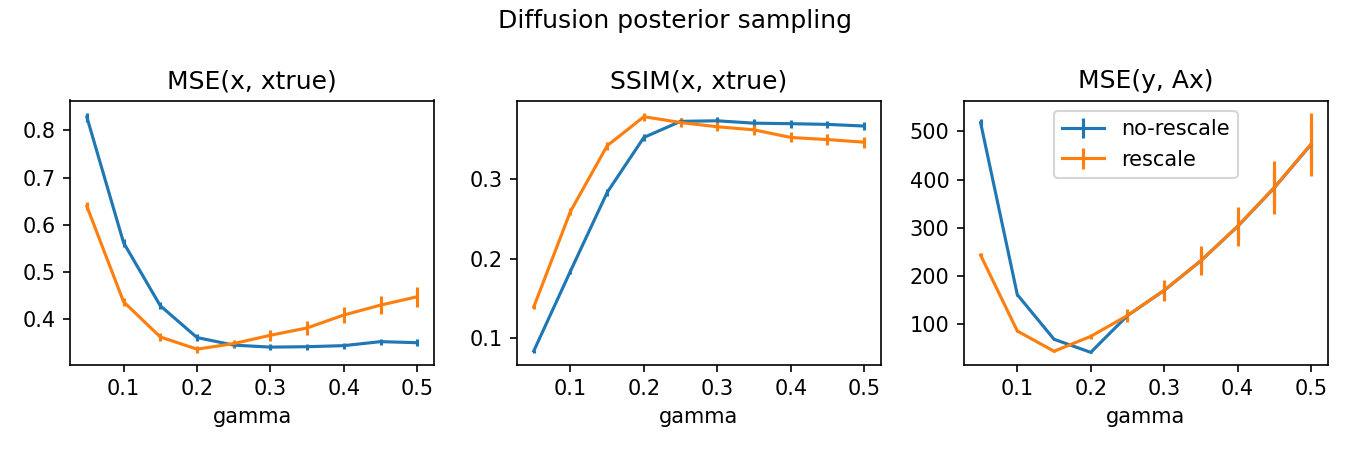}
    \caption{Varying $\gamma$ (from Eq. 5 and 6 of the main text) and comparing the unscaled and rescaled versions of Diffusion Posterior Sampling. We keep the diffusion schedule from the paper that we found gave high quality unconditional generations. This schedule is described in Eq. 2 of the main text, but in brief we use 50 iterations with an exponentially decaying noise schedule.  }
    \label{fig:my_label}
\end{figure}

\begin{figure}[H]
    \centering
    \includegraphics[width=\linewidth]{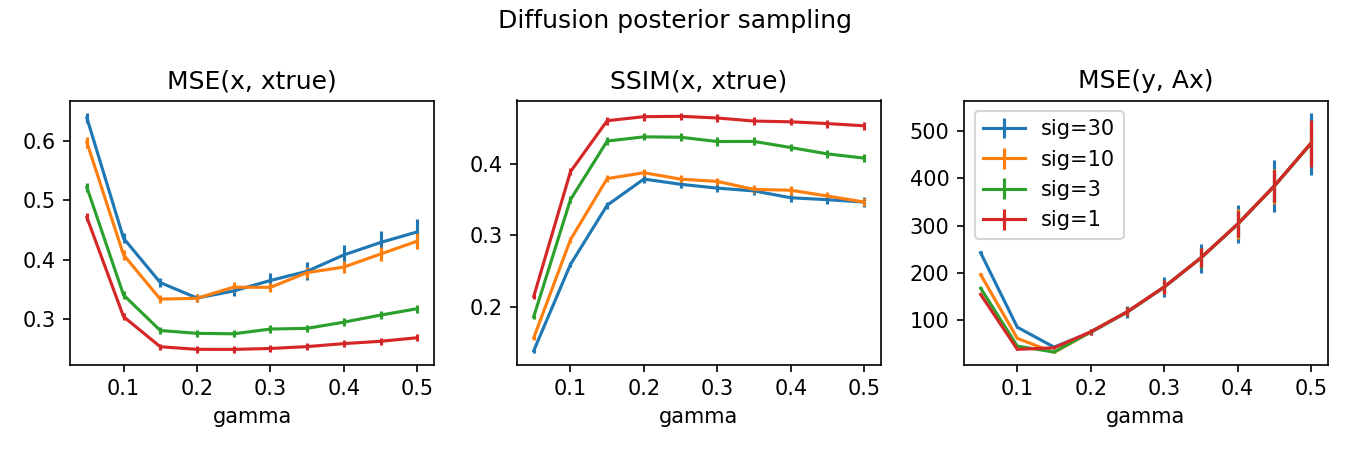}
    \caption{We were able to improve performance of DPS by modifying the diffusion noise schedule. In particular, we try different starting noise level $\sigma_1$, and explore performance of the rescaled version of DPS for various $\gamma$. Note that we choose our schedule according to $\sigma_n = \sigma_1 (\sigma_N/\sigma_1)^{(n-1)/(N-1)}$ so that we keep the number of diffusion steps fixed when we decrease the starting noise level (the spacing between noise levels just decreases as we decrease the starting noise level). In this figure, we still use $50$ total denoiser evaluations.}
    \label{fig:my_label}
\end{figure}

\begin{figure}[H]
    \centering
    \includegraphics[width=\linewidth]{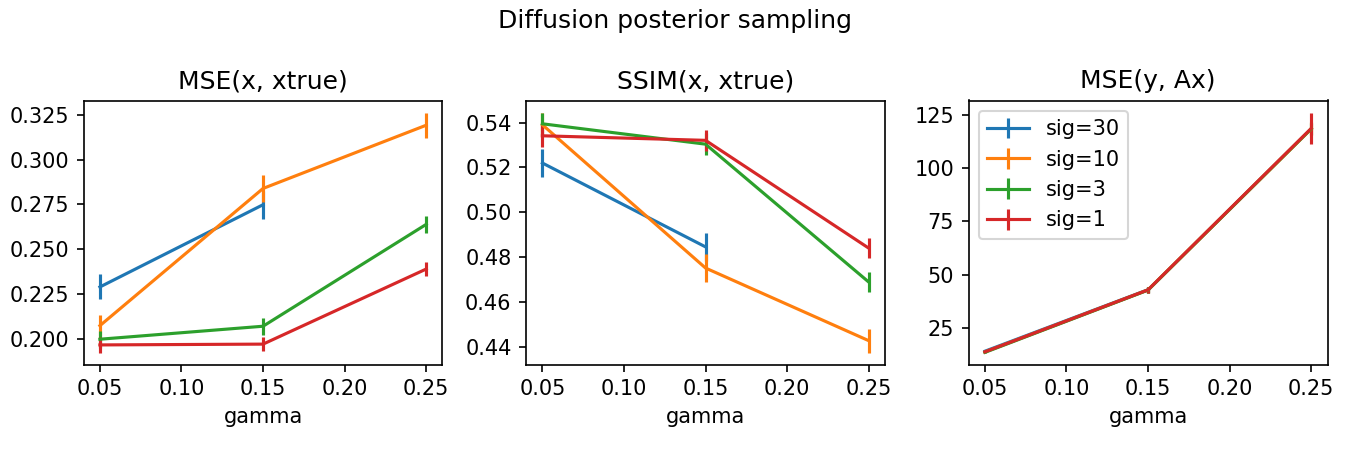}
    \caption{We evaluate DPS now with 1000 diffusion iterations. We choose our schedule according to $\sigma_n = \sigma_1 (\sigma_N/\sigma_1)^{(n-1)/(N-1)}$. We vary $\gamma$ and the starting $\sigma_1$. Interestingly the data error terms (right plot) $Ax -y$ are nearly identical for all configurations. }
    \label{fig:dps_niter1000}
\end{figure}

\section{Arbitrary sized reconstruction}
We modify the noise-prediction network itself to run in patches, then we smoothly average them together. This is the Approach diagrammed in Figure 1 of \cite{wu2021chunkflow}. Mathematically we run:
\begin{equation}
    \epsilon_{\text{patchified}}(x)[i,j] = \frac{\sum_{u,v=1}^{\infty} B[i-su,j-sv] \epsilon_\theta(x[su:su+p,sv:sv+p])[i-su,j-sv]} {\sum_{u,v} B[i-su,j-sv] }
\end{equation}
where $s=96$ is the stride, and $p=128$ is the patch size, that we use in the experiments. The bump function we use is the product of two 1D bump functions
\begin{equation}
    B[x, y] = b(2x/p-1) \cdot b(2y/p-1)
\end{equation}
and each 1D bump function is given by:
\begin{equation}
    b(u) = 
    \begin{cases}
        1-\exp\left(\frac{-1}{\max\{1-u^2,0.2\}}\right) &\text{ if } |u| < 1 \\
        0 &\text{otherwise}
    \end{cases}
\end{equation}
These decay smoothly to $0.2$ as $x\rightarrow +p/2$ and $x \rightarrow -p/2$. With this overlap fraction, pixels are on averaged processed $1.8\times$ by the network so this method is approximately $1.8x$ slower than just running a larger patch through our network.


\end{document}